\documentclass[journal]{IEEEtran}

\usepackage[colorlinks,urlcolor=blue,linkcolor=blue,citecolor=blue]{hyperref}
\usepackage{color,array}
\usepackage{graphicx}
\usepackage{amsmath,amssymb}
\usepackage{stfloats}
\usepackage[caption=false,font=normalsize,labelfont=sf,textfont=sf]{subfig}
\usepackage{algorithm}
\usepackage{algpseudocode}

\newcommand{\RomanNumeralCaps}[1]
    {\MakeUppercase{\romannumeral #1}}

\begin{document}

\title{Credit-cognisant reinforcement learning for multi-agent cooperation}

\author{F. Bredell, H. A. Engelbrecht, \IEEEmembership{Senior Member, IEEE}, and J. C. Schoeman, \IEEEmembership{Member, IEEE} \thanks{This paper was first submitted for review on
    22 October 2022. This work was supported in part by grants from the DW Ackerman PG Bursary as administered by Stellenbsoch University from 2021 to 2022.} \thanks{ The authors are with
    the Electrical and Electronic Department of Stellenbsoch University, Stellenbosch, 7600, South Africa (e-mail: 20795718@sun.ac.za (F. Bredell); hebrecht@sun.ac.za (Prof. H. A.
    Engelbrecht); jcschoeman@sun.ac.za (Dr. J. C. Schoeman)).}}

\markboth{Journal of IEEE Transactions on Games, Vol. ??, No. ?, ?? 2022}
{F. Bredell \MakeLowercase{\textit{et al.}}: CCRs for multi-agent cooperation}

\maketitle

\begin{abstract}
    Traditional multi-agent reinforcement learning (MARL) algorithms, such as independent Q-learning, struggle when presented with partially observable scenarios, and where agents
    are required to develop delicate action sequences. This is often the result of the reward for a good action only being available after other agent's have taken theirs, and
    these actions are not credited accordingly. Recurrent neural networks have proven to be a viable solution strategy for solving these types of problems, resulting in significant
    performance increase when compared to other methods. In this paper, we explore a different approach and focus on the experiences used to update the action-value functions of
    each agent. We introduce the concept of credit-cognisant rewards (CCRs), which allows an agent to perceive the effect its actions had on the environment as well as on its
    co-agents. We show that by manipulating these experiences and constructing the reward contained within them to include the rewards received by \textit{all} the agents within
    the same action sequence, we are able to improve significantly on the performance of independent deep Q-learning as well as deep recurrent Q-learning. We evaluate and test the
    performance of CCRs when applied to deep reinforcement learning techniques at the hands of a simplified version of the popular card game \textit{Hanabi}. 

\end{abstract}


\section{Introduction}
    \IEEEPARstart{R}{einforcement} learning (RL) holds promising potential to address a large variety of problems where artificial agent operation offers a significant improvement
    over alternative methods~\cite{marl_why_good}. For certain real-world problems, single-agent operation is not optimal (or even possible) and the incorporation of multiple
    agents would be beneficial (or necessary). Scenarios such as autonomous vehicles navigating terrain~\cite{automated_MARL_vehicles}, guided drone
    swarms~\cite{guided_drone_swarms}, mapping verbal instructions to executable actions~\cite{map_instr_to_act}, or switching of railway lines benefit from the use of multiple
    agents. RL could help agents to effectively cooperate within these multi-agent scenarios by learning from past and or simulated experiences. 

    Unfortunately, the introduction of multiple agents into an environment typically increases the complexity of the problem exponentially. It introduces a moving target
    learning problem~\cite{MARL_survey}, since all the agents must learn simultaneously. The reason for this is that each individual agent's policy changes over time, which in turn
    causes a non-stationary environment. This could inhibit all the agents from developing effective policies and produce undesired behaviour~\cite{MARL_survey}. 
    
    One of the most straightforward solutions to multi-agent reinforcement learning (MARL) is the combination of deep Q-networks (DQNs) and independent Q-learning~\cite{idqn_pong},
    where each agent independently and simultaneously learns its own action-value function (or \textit{Q-values}) while interacting with the same environment. However, this
    strategy does not fair well when paired with partial observability\footnote{Partial observability refers to problems where the full state space is hidden from each individual
    agent, resulting in each agent having their own unique perspective of the problem.}~\cite{iql_short_og, iql_short_new}. Hausknecht and Stone~\cite{drqn_stone} have shown
    that recurrent neural networks offer an improved solution to MARL problems containing partial observability. These networks incorporate a built-in short term memory over which
    experiences are \emph{unrolled} (or combined to form longer sequences). This is often combined with deep Q-learning's feed forward neural network to produce deep recurrent
    Q-learning~\cite{drqn_stone}. 

    In this paper, we specifically focus on turn-based MARL problems containing partial observability, and where a \textit{delicate action sequence} is required for successful
    cooperation. We define a delicate action sequence as a set of consecutive actions, where each action is imperative and highly correlated to the others. Changing any one action
    in such a sequence would make the difference between receiving a reward or not, and the agents will not be able to effectively solve the problem at hand. Most research efforts
    explore the implementation of advanced algorithms with powerful function approximators to solve similar MARL problems~\cite{foerster_rial_dial, rashid2018qmix}. 
    
    A popular approach is to use games as a medium for testing and evaluating RL algorithms, where examples include \textit{Go}~\cite{silver2016alphago},
    \textit{Backgammon}~\cite{gammon}, and \textit{Dota 2}~\cite{openai_five}. In this paper, we use a simplified version of the popular card game \textit{Hanabi}, which has
    recently become an area of interest for testing and developing MARL algorithms~\cite{hanabi_ai}, as an illustrative example. 
    
\subsection{Related work}\label{related}  
    Sunehag \textit{et al.}~\cite{vdns} propose an alternative solution to independent deep Q-learning called value decomposition networks (VDNs). Instead of each agent learning
    independently, the agents learn a joint action-value function which is equal to a linear combination of each agent's individual action-value functions~\cite{vdns}. VDNs allow
    for the automatic decomposition of complex learning problems into local simplified sub-problems for each agent. However, VDNs can only represent a limited class of joint
    action-value functions and do not scale well when increasing the number of agents~\cite{rashid2018qmix, vdns}.

    Rashid \textit{et al.}~\cite{rashid2018qmix} introduce an extension of VDNs called \textit{QMIX}, allowing agents to cooperate in a more complex environment. They explore and
    evaluate this method within the context of the video game \textit{StarCraft \RomanNumeralCaps{2}}. QMIX uses a combination of recurrent neural networks and hypernetworks to act
    as a \textit{non-linear} combination of each agent's action-value functions to achieve remarkable performance in this environment. Although remarkable, this approach has high
    complexity since it involves the combination of multiple networks and a non-linear combination of action-value functions, which can lead to sub-optimal
    polices~\cite{qmix_shortcomings}. 

    To achieve effective communication between agents within a partially observable environment, Foerster \textit{et al.}~\cite{foerster_rial_dial} explore the capabilities of
    multiple deep RL agents and develop two methods called \textit{Reinforced Inter-Agent Learning (RIAL)} and \textit{Differentiable Inter-Agent Learning (DIAL)}. These agents can
    directly share weights of their networks via communication, to allow for the development of cooperative policies. Although simpler than QMIX, their research is specifically
    focused on the communication protocol developed between agents and how they can cooperate effectively through communication. 

    Another alternative is the distribution of the joint action-value function between agents, proposed by Schneider \textit{et al.}~\cite{1999_distributed_val}. This distribution
    is based on each agent's predefined contribution within the global reward function. Guestrin \textit{et al.}~\cite{coordination_graphs} further build on this research by
    introducing \textit{coordination graphs} that allow for the exploitation of conditional independencies between agents. This approach allows the agents to coordinate their
    actions more effectively. However, in both cases this requires the agent's dependencies on each other to be predefined and their influence within the global reward to be known.

    \textit{Rainbow}, introduced by Hessel \textit{et al.}~\cite{rainbow}, offers a state-of-the-art solution. It combines various advancements made to deep Q-learning, and has
    proven to offer significant performance gain when faced with large discrete action spaces~\cite{rainbow_energy}. A natural extension of Rainbow to multi-agent systems, is the
    implementation of independent agents (referred to as MARainbow)~\cite{marainbow}. Bard \textit{et al.}~\cite{hanabi_ai} explore the capabilities of MARainbow within Hanabi as a
    preliminary solution strategy. However, Rainbow introduces a plethora of new hyperparameters which can often result in sub-optimal policies~\cite{rainbow_shortcoming}.  

\subsection{Summary of Contributions}
    We argue that delicate action sequences pose a challenge to the traditional RL update step. We will prove this by implementing tabular RL within a simplified problem, to
    illustrate the shortcoming of independent Q-learning. To achieve this, we extend existing RL methods to turn-based, partially observable environments using our unifying
    notation. Using this extension, we introduce our novel contribution, called credit-cognisant rewards (CCRs), and show how it improves on existing RL methods. 

    The key to CCR is the fact that the reward received for a good action is usually only available after other agents have taken their actions, and that these actions are not
    credited accordingly. We contrast this to distributed value functions and coordination graphs where the dependencies between agents and their role within the global reward is
    predefined. We expect our agents to learn these dependencies without requiring predefined knowledge. This is also in contrast to RIAL and DIAL, where agents share network
    weights directly, and thereby increasing the problem complexity by introducing an additional communication channel between agents. Additionally, our approach does not introduce
    any new hyperparameters, and acts as a variation to existing methods utilising traditional transition tuple. 

    Our approach also differs from n-step bootstrapping, which shares similarities, but implements a different concept. We also contrast this to recurrent neural networks where the
    focus is on architecture design and function approximation, whereas our focus is a fundamental modification of the experiences received by each agent. We compare and evaluate
    independent Q-learning, independent deep Q-learning and deep recurrent Q-learning along with their CCR variants, to demonstrate the shortcomings of existing methods and
    illustrate how CCR can lead to improved performance.

\subsection{Paper Outline}
    To discuss some of the core concepts of RL, we first consider the single agent system with full observability in Section~\ref{prelim}, and introduce deep reinforcement learning
    with standard terminology. We then extend these concepts to turn-based MARL containing partial observability with our unifying notation in Section~\ref{extend}. This is
    followed by the introduction and formal definition of credit-cognisant rewards (CCRs) in Section~\ref{CCR}. We then introduce our example settings, namely a tabular problem and
    \emph{colourless Hanabi}, in Section~\ref{setting_1} and~\ref{setting_2}. For each example, we discuss the experimental setup for each algorithm, and present and discuss our results.

\section{Preliminaries}\label{prelim} 
    Reinforcement learning allows agents to develop policies based on past experiences~\cite{sutton_barto}. These policies are usually aimed at maximizing a reward $R$ received
    from the environment after taking action $A$ within a given state $S$. This results in the transition tuple
    \begin{equation}\label{classic_sarsa}
        e = \langle S_t, A_t, R_{t+1}, S_{t+1} \rangle
    \end{equation}
    at timestep $t$~\cite{sutton_barto}. These transition tuples are used to calculate the value functions which acts as a quantitative measure for the desirability of a state. In
    tabular methods these values are represented using a lookup-table, while deep RL methods use powerful function approximators to estimate these values. 

\subsection{Independent Q-learning}
    Q-learning is an off-policy, temporal difference (TD) control algorithm that learns the action-value function $Q(S,A)$ by directly approximating the optimal action-value
    function $Q^*(S,A)$, independent of the policy being followed~\cite{sutton_barto}. The update step for the action-value function is defined as
    \begin{equation}\label{update_step_ql}
        \begin{aligned}
            Q(S_t, A_t) \gets {} & Q(S_t, A_t) + \alpha \bigr[ R_{t+1} + \\
                                        & \gamma \max_{a} Q(S_{t+1}, a) - Q(S_t, A_t) \bigr],
        \end{aligned}
    \end{equation}
    where $\alpha$ is the learning rate and $\gamma$ the discount factor of future rewards~\cite{sutton_barto}. 
    
    In Q-learning the agent selects its actions based on an $\epsilon$-greedy strategy, i.e., 
    \begin{equation}
        a =  \begin{cases}
            \text{argmax}_a Q(S_t, a) & \text{with probability}~1 - \epsilon    \\
            \text{random action}    &   \text{with probability}~\epsilon
        \end{cases},
    \end{equation}
    where $\epsilon$ is the exploration rate to account for exploration of the state-space~\cite{sutton_barto}. This can be extended to MARL with independent Q-learning, where each
    agent has their own action-value function and uses their individual experiences in their update steps~\cite{iql_og}.   

    Instead of distinct policies, independent Q-learning agents can make use of a shared policy, especially when the environment contains symmetries~\cite{iql_og}. This allows the
    update step (\ref{update_step_ql}) to be updated more frequently by using the experiences of all the independent agents. This has proven to offer significant performance gain
    and allow agents to learn more effectively~\cite{iql_og}.

\subsection{N-step Bootstrapping}
    N-step bootstrapping is a method unifying one-step TD learning and Monte Carlo methods, to allow an agent's action-values to be bootstrapped over a longer time
    sequence~\cite{sutton_barto}. In n-step methods, the rewards are constructed over $n$ actions, resulting in the update step 
    \begin{equation}\label{update_step_n_step}
        \begin{aligned}
            Q(S_t, A_t) \gets {} & Q(S_t, A_t) + \alpha \biggr[ \sum_{i=0}^{n-1} \gamma^i R_{t+1+i} \\
                                        & + \gamma^n \max_{a} Q \bigr( S_{t+n}, a \bigr) - Q(S_t, A_t) \biggr].
        \end{aligned}
    \end{equation}
    For $n = 1$, (\ref{update_step_n_step}) simplifies to (\ref{update_step_ql}), and if $n = \infty$ the update step is calculated over the entire episode of experiences
    (identical to Monte Carlo methods). It is important to note that the update step in (\ref{update_step_n_step}) can only occur at timestep $t+n$, to ensure that all the rewards
    are available. This allows an agent to develop improved policies based on more than just one experience, without waiting for the episode to terminate. We will later discuss how
    this differs from our new variant. 

\subsection{Deep Q-learning}\label{DQN} 
    Deep Q-learning is an extension of tabular Q-learning where artificial neural networks (referred to as deep Q-networks or DQNs) are used as non-linear function
    approximators~\cite{dqn_main}. Unfortunately, RL often becomes unstable or diverges from the desired policy when using non-linear function approximators~\cite{dqn_main}. To
    solve this instability, deep Q-learning implements an experience replay memory to store the experiences $e$~\cite{replay_mem}. These experiences are usually sampled in random
    batches $b$ to remove correlation within the observation sequence, thereby smoothing over the changes within the data distribution~\cite{dqn_main}. 
    
    Deep Q-learning usually incorporates a \textit{policy network} and a \textit{target network}~\cite{dqn_main}. The policy network is used to select actions and utilises the
    update step, while the target network serves as a baseline when calculating the \textit{TD error}. The target network is updated periodically with the policy network to reduce
    correlation with the target. 
    
    The weights of the policy network are updated using backpropagation with the goal of minimizing the TD error, which is defined as
    \begin{equation}\label{td error}
        \begin{aligned}
            L_i(\theta_i^{}) = {}  & \mathbb{E}_{b}[(R_{t+1} + \gamma \max_{a}Q(S_{t+1}, a;\theta_i^-) \\
                                & - Q(S_t, A_t; \theta_i^{}))^2],
        \end{aligned}
    \end{equation}
    where $\theta_i^{}$ and $\theta_i^-$ are the weights of the policy-and target network, respectively, at iteration $i$~\cite{dqn_main}. To obtain the update step for the
    weights of the policy network, the TD error defined in (\ref{td error}) must be differentiated according to the weights to obtain
    \begin{equation}\label{td update}
        \begin{aligned}
            \theta_{i+1}^{} = {}  & \theta_i^{} + \alpha [R_{t+1} + \gamma \max_{a}Q(S_{t+1}, a;\theta_i^-) \\
                                & - Q(S_t, A_t; \theta_i^{})]\nabla_{\theta_i}Q(S_t, A_t; \theta_i^{}),
        \end{aligned}
    \end{equation}
    which Sutton and Barto~\cite{sutton_barto} refer to as the semi-gradient form of Q-learning~\cite{dqn_main}. In practice the backpropagation and calculation of the TD error is
    usually handled by an optimizer, such as the \textit{Adam} optimizer~\cite{adam_opt}.

    Similar to tabular Q-learning, deep Q-learning can be extended to MARL using independent deep Q-learning~\cite{idqn_pong}. However, due to the addition of DQNs and the added
    complexity of the moving target learning problem, this approach does not have any guarantees of convergence. In spite of this uncertainty, it has a strong track record and has
    shown some remarkable performance in the game of \textit{Pong}~\cite{idqn_pong}, and is often considered a baseline when measuring the performance of other algorithms.  

\subsection{Deep Recurrent Q-Learning}\label{DRQN} 
    In deep Q-learning, the algorithm performs better with full observability of the environment. However, for MARL problems with partial observability, deep recurrent Q-learning
    has shown to be a superior alternative~\cite{drqn_stone}. Instead of approximating the action-value function with a fully-connected feed forward neural network, deep recurrent
    Q-learning adds recurrent layers to the network. These layers can maintain a history of visited states and accumulate observations over time, acting as a built-in memory
    system~\cite{drqn_stone}. 
    
    These recurrent layers are typically gated architectures such as long short-term memory (LSTM)~\cite{lstm} or gated recurrent unit (GRU)~\cite{gru} which allows learning over
    longer timescales. It introduces a new hyperparameter called the unrolled length $K$, which determines the number of iteration timesteps over which the update step is
    calculated. This offers remarkable performance gain, and will be considered as a viable alternative solution strategy during our evaluation. 

    \section{Modelling Partial Observability}\label{extend}
    We now extend independent Q-learning, n-step bootstrapping, and deep Q-learning to include turn-based environments containing partial observability using our unifying
    notation. This will allow us to introduce credit-cognisant rewards (CCRs) and show how it builds on these concepts. 

\subsection{Transition Tuple}
    First, we must change the transition tuple in (\ref{classic_sarsa}) to account for multiple agents by separating their individual states based on their perspectives. These
    perspectives will be unique due to the partial observability of the problems we consider. Therefore, we can separate the perspectives of each agent $i$ using
    \begin{equation}\label{pp}
        i = t - rP,
    \end{equation}
    where $P$ is the total number of players within the environment and $r$ the current round.

    Since the problems we consider are turn-based, each action and reward at timestep $t$ will be associated with agent $i$ within that round, and does not require discerning based
    on perspectives. Using this we can extend the transition tuple in (\ref{classic_sarsa}) to a turn-based multi-agent system with
    \begin{equation}\label{multi_agent_sarsa}
        e = \langle S^i_t, A_t, R_{t+1}, S^i_{t+1} \rangle.
    \end{equation}
    These tuples will be used to update the action-value functions of each agent within the MARL environment. Table~\ref{i_table} illustrates a typical agent experience when
    interacting with a turn-based environment containing three agents.

    \begin{table} [h]
        \caption{Summary of notation for a three agent environment.}\label{i_table}
        \setlength{\tabcolsep}{3pt}
        \begin{tabular*}{21pc}{@{}|p{25pt}|p{30pt}<{\raggedright\hangindent6pt}|p{30pt}<{\raggedright\hangindent6pt}|p{25pt}<{\raggedright\hangindent6pt}|p{25pt}<{\raggedright\hangindent6pt}|p{30pt}<{\raggedright\hangindent6pt}|p{40pt}<{\raggedright\hangindent6pt}|@{}}
            \hline
            \textbf{Time(t)}    &   \textbf{Round(r)}  &   \textbf{Agent(i)}  &   \textbf{State}   &   \textbf{Action}    &   \textbf{Reward}     &   \textbf{New State} \\
            \hline
            0   &   0   &   0   &   $S^0_0$ &   $A_0$   &   $R_1$   &   $S^0_1$  \\
            1   &   0   &   1   &   $S^1_1$ &   $A_1$   &   $R_2$   &   $S^1_2$  \\
            2   &   0   &   2   &   $S^2_2$ &   $A_2$   &   $R_3$   &   $S^2_3$  \\
            3   &   1   &   0   &   $S^0_3$ &   $A_3$   &   $R_4$   &   $S^0_4$  \\
            4   &   1   &   1   &   $S^1_4$ &   $A_4$   &   $R_5$   &   $S^1_5$  \\
            5   &   1   &   2   &   $S^2_5$ &   $A_5$   &   $R_6$   &   $S^2_6$  \\
            6   &   2   &   0   &   $S^0_6$ &   $A_6$   &   $R_7$   &   $S^0_7$  \\
                &       &       &   \dots   &           &           &            \\
            t   &   r   &   $i=t-rP$   &   $S^i_t$ &   $A_t$   &   $R_{t+1}$   &   $S^i_{t+1}$  \\     
            \hline
            \multicolumn{7}{@{}p{21pc}@{}}{}\\
        \end{tabular*}\label{tab1}
    \end{table}

\subsection{Independent Q-learning and Deep Q-learning}
    Using these new tuples, we can extend the update step for independent Q-learning (\ref{update_step_ql}) to include partial observability with
    \begin{equation}\label{update_step_iql}
        \begin{aligned}
            Q(S^i_t, A_t) \gets {} & Q(S^i_t, A_t) + \alpha \bigr[ R_{t+1} + \\
                                        & \gamma \max_{a} Q(S^i_{t+1}, a) - Q(S^i_t, A_t) \bigr].
        \end{aligned}
    \end{equation}
    This ensures that each agent's update step is constructed over their own experiences, but similar to (\ref{update_step_ql}) the agents can utilise a shared policy to increase
    the overall performance. This results in the algorithm shown in Algorithm~\ref{ql algorithm}, where the agents implement an $\epsilon$-greedy policy. The experiences resulting
    from lines 7 and 8 will follow the same structure as the transition tuples illustrated in Table~\ref{i_table}.

    \begin{algorithm}
        \caption{Independent Q-learning in a turn-based environment}\label{ql algorithm}
        \begin{algorithmic}[1]
            \State Initialise hyperparameters: learning rate $\alpha \in (\,0, 1]\,$, discount factor $\gamma \in (\,0, 1]\,$, total number of player $P$, and exploration rate $\epsilon \in (\,0, 1)\,$
            \State Initialise all $Q(S, A) \gets 0$

            \For{each episode}
                \State Reset environment and set $r \gets 0$ and $t \gets 0$

                \While{$S_{t+1}^{0:P-1}$ is not terminal}
                    \For{all players $i$ from $0$ to $P-1$}
                        \State Choose $A_t$ from $S^i_t$ using the policy derived from $Q$
                        \State Take action $A_t$ and observe $R_{t+1}^{}$, $S^i_{t+1}$
                        \State $Q(S^i_t, A_t) \gets Q(S^i_t, A_t) + \alpha \bigr[ R_{t+1} + \gamma \max_{a} Q(S^i_{t+1}, a) - Q(S^i_t, A_t) \bigr]$
                        \State $t \gets t + 1$
                    \EndFor
                    \State $r \gets r+1$
                \EndWhile
            \EndFor
        \end{algorithmic}
    \end{algorithm}  

    Similarly, we can extend the update step for deep Q-learning (\ref{td update}) using the tuples in (\ref{multi_agent_sarsa}) to obtain
    \begin{equation}\label{td update multi}
        \begin{aligned}
            \theta_{i+1}^{} = {}  & \theta_i^{} + \alpha [R_{t+1} + \gamma \max_{a}Q(S_{t+1}^i, a;\theta_i^-) \\
                                & - Q(S_t^i, A_t; \theta_i^{})]\nabla_{\theta_i}Q(S_t^i, A_t; \theta_i^{}).
        \end{aligned}
    \end{equation}
    This ensures that the policy network is updated using the states and actions of an individual agent based on its perspective. It will follow a similar strategy as
    Algorithm~\ref{ql algorithm}, with the calculation of the TD error and network weights' backpropagation replacing the update step in line 9.  

\subsection{Independent N-step Bootstrapping}
    We can also extend n-step bootstrapping for turn-based environments containing partial observability using the new transition tuples in (\ref{multi_agent_sarsa}). This results
    in the update step 
    \begin{equation}\label{update_step_n_step_multi}
        \begin{aligned}
            Q(S^i_t, A_t) \gets {} & Q(S^i_t, A_t) + \alpha \biggr[ \sum_{k=0}^{n-1} \gamma^k R_{t+1+kP} \\
                                        & + \gamma^n \max_{a} Q \bigr( S^i_{t+1+(n-1)P}, a \bigr) \\
                                        & - Q(S^i_t, A_t) \biggr].
        \end{aligned}
    \end{equation}
    Similar to the fully observable case, this requires the update step to be delayed until all the rewards are available. However, since the rewards are combined over an
    \textit{individual} agent's experiences, these rewards must be delayed by a timestep $t+n$ as well as the other agent's turns within the episode, i.e., $t+1+(n-1)P$. 

\section{Credit-cognisant Rewards}\label{CCR} 
    Traditional MARL methods struggle when presented with partial observability, where a delicate action sequence is required for effective cooperation. To overcome this challenge,
    we introduce a novel approach that incorporates the rewards received by all the agents within a given action sequence. For this purpose, we define the credit-cognisant reward
    (CCR) as
    \begin{equation}
        C_{t+1} = \sum_{k=0}^{P-1} R_{t+1+k}.
    \end{equation}
    
    This allows an agent to perceive the effect their action had on the environment as well as on the other agents contained within it, and encourages the development of
    improved action sequences. It acts as an action-reaction mechanism where an agent observes their immediate reward for taking an action, and incorporates the consecutive agents'
    rewards into this immediate reward as a reaction to the first agent's action. 
    
\subsection{CCR Transition Tuple}
    We can, therefore, change the multi-agent transition tuple defined in (\ref{multi_agent_sarsa}) to include the CCR as
    \begin{equation}\label{multi_agent_ccr_sarsa}
        e = \langle S_t^i, A_t, C_{t+1}, S_{t+P}^i \rangle.
    \end{equation}
    The next state $S_{t+1}^i$, typically used in the estimation value, has changed to $S_{t+P}^i$, i.e., the new state as seen from the current agent's perspective after the other
    agents have taken their turns. This allows the other agent's actions to be seen as part of the `environment step' resulting from action $A_t$, effectively mimicking a
    single-agent environment interaction. 
    
    It is important to note that $C_{t+1}$ is only available after a timestep of $t+P$ (or one round), resulting in a delay when updating the action-value functions (similar to
    n-step). 

\subsection{CCR applied to Independent Q-learning}
    To apply CCRs to independent Q-learning, we must incorporate the new transition tuple (\ref{multi_agent_ccr_sarsa}) into the update step (\ref{update_step_iql}) of the
    action-value function for each agent, resulting in 
    \begin{equation}\label{update_step_ccr}
        \begin{aligned}
            Q(S_t^i, A_t) \gets {} & Q(S_t^i, A_t) + \alpha \biggr[ C_{t+1} \\
                                        & + \gamma \max_{a} Q \bigr( S_{t+P}^i, a \bigr) - Q(S_t^i, A_t) \biggr]
        \end{aligned}
    \end{equation}

    This requires that we store each agent's action and reward within a temporary buffer for each action sequence, to allow the calculation of the CCR for each action after a delay
    of $P$. This will result in a delay within the update step of line 9 in the independent Q-learning method found in Algorithm~\ref{ql algorithm}, while also introducing the
    temporary buffer in line 8. Therefore, we now update the algorithm found in Algorithm~\ref{ql algorithm} to produce Algorithm~\ref{ccr algorithm}, which acts as the algorithm
    for independent Q-learning with CCR applied. In this algorithm, $\tau$ represents the time step lagging the current time step $t$ by the number of players $P$.  

    \begin{algorithm}
        \caption{Independent Q-learning with CCRs applied}\label{ccr algorithm}
        \begin{algorithmic}[1]
            \State Initialisation similar to Algorithm~\ref{ql algorithm}

            \For{each episode}
                \State Reset environment and set $r \gets 0$ and $t \gets 0$

                \While{$S_{t+1}^{0:P-1}$ is not terminal}
                    \For{all players $i$ from $0$ to $P-1$}
                        \State Choose $A_t$ from $S_t^i$ using the policy derived from $Q$
                        \State Take action $A_t$ and observe $R_{t+1}^{}$, $S_{t+1}^{0:P-1}$

                        \If{$r \ge 1$}
                            \State $\tau \gets t - P$
                            \State $C_{\tau+1} = \sum_{k=0}^{P-1} R_{\tau+1+k}$
                            \State $Q(S_\tau^i, A_\tau) \gets Q(S_\tau^i, A_\tau) + \alpha [C_{\tau+1} + \gamma \max_{a} Q(S_{\tau+P}^i, a) - Q(S_\tau^i, A_\tau)]$
                        \EndIf

                        \State $t \gets t + 1$
                    \EndFor

                    \State $r \gets r + 1$
                \EndWhile

                \For{$j=i+1$ until $t - rP$}
                    \State $\tau \gets \tau + 1$
                    \State $C_{\tau+1} = \sum_{k=0}^{P-1} R_{\tau+1+k}$
                    \State $Q(S_\tau^j, A_\tau) \gets Q(S_\tau^j, A_\tau) + \alpha [C_{\tau+1} + \gamma \max_{a} Q(S_{t+1}^j, a) - Q(S_\tau^j, A_\tau)]$
                \EndFor

            \EndFor
        \end{algorithmic}
    \end{algorithm}

    Similar to Algorithm~\ref{ql algorithm}, the first few lines act as initialisation of hyperparameters and is identical. When the agents select and execute actions in line 6 and
    7, the rewards and new states must be stored within the temporary buffer. As mentioned previously, the calculation of the CCR can only occur after 1 round, and is incorporated
    into line 8. Line 9 illustrates the lagging time step $\tau$, which allows for the calculation of the CCRs and updating of the value functions for the current player $i$ using
    the rewards of the previous action sequence. 

    Importantly, lines 17 to 21 illustrate the terminal round calculation, which is considered a special case. The terminal round calculation requires that an agent's final rewards
    are calculated based on the rewards received after it took its final action until the terminal state. Additionally, it requires that the update step includes the terminal state as seen by
    all the agents from their individual perspectives ($S_{t+1}^{0:P-1}$). 

    We note that CCR requires the sharing of rewards during training, which in practice is not always feasible. However, the agents can utilise centralized training with
    decentralized execution allowing for the sharing of reward information in the training phase~\cite{central_decentral}. During execution the agents act independently without
    sharing additional information, and act greedily according to their individual action-value functions. 
    
    Furthermore, the delay imposed in the update step is only present when training the agents, during execution there is no difference between the original method and its CCR
    variant. Similar to this discussion, the CCR variant can be extended to any method utilising the traditional transition tuple. 

\subsection{CCR vs N-step Bootstrapping}
    CCR shares similarities with n-step bootstrapping, but in fact utilises a different concept. The update step defined in (\ref{update_step_ccr}) illustrates the major
    differences between n-step bootstrapping and CCRs. The n-step rewards are constructed over $n$ actions of an individual agent's experience, e.g., if $P=3$ and $n=2$, then agent
    0 will calculate its rewards based on the rewards received for actions $A_0$ and $A_3$ (see Table~\ref{i_table}).
    
    On the contrary, the CCR variant constructs an agent's reward based on the rewards received by all the agents within the same action sequence, e.g., for the same setup where
    $P=3$, agent 0 will calculate the reward based on the rewards received for actions $A_0$, $A_1$ and $A_2$ (see Table~\ref{i_table}). This allows the agent to learn from rewards
    that were received on another agent's turn as an indirect consequence of its action, as oppose to its own rewards over multiple actions. 
    
    Additionally, CCR does not introduce a new hyperparameter (unlike n-step bootstrapping) and only acts as a variation to any MARL method using traditional transition tuples.
    Another important difference between n-step bootstrapping and CCRs is the next state used in the update step. For n-step bootstrapping the next state is $S_{t+1+(n-1)P}^i$, as
    seen in (\ref{update_step_n_step_multi}), while CCR uses $S_{t+P}^i$. This reduces the severe shift in an agent's perspective as the environment transitions from one state to
    the next as a result of each agent taking an action, e.g., the difference between $S_4^0$ and $S_3^0$ for agent 0 in our example mentioned previously. During training and
    evaluation we will compare CCRs to n-step bootstrapping to further illustrate the difference between these two methods. 

\subsection{CCR applied to Deep Reinforcement Learning}
    Similar to tabular methods, when applying CCR to deep RL it only affects the experiences stored within the replay memory of each agent. In the case of deep Q-learning, these
    experiences will consist of tuples identical to the ones defined in (\ref{multi_agent_ccr_sarsa}). These experiences are then sampled in random batches, which is used to
    calculate the new TD error 
    \begin{equation}\label{td error ccr}
        \begin{aligned}
            L_i(\theta_i^{}) = {}  & \mathbb{E}_{b} \biggr[ \bigr( C_{t+1} + \gamma \max_{a} Q \bigr( S_{t+P}^i, a;\theta_i^- \bigr) \\
                                & - Q(S_t^i, A_t; \theta_i^{}) \bigr) ^2 \biggr].
        \end{aligned}
    \end{equation}
    
    Similar to the TD error defined in (\ref{td error}), we can differentiate according to the weights of the policy network to obtain the semi-gradient form of deep Q-learning
    with CCR applied,
    \begin{equation}\label{td update ccr}
        \begin{aligned}
            \theta_{i+1}^{} = {}  & \theta_i^{} + \alpha \biggr[ C_{t+1} + \gamma \max_{a} Q \bigr( S_{t+P}^i, a;\theta_i^- \bigr) \\
                                & - Q(S_t^i, A_t; \theta_i^{}) \biggr] \nabla_{\theta_i}Q(S_t^i, A_t; \theta_i^{}).
        \end{aligned}
    \end{equation}
    Deep Q-learning with CCR applied will follow a similar structure as Algorithm~\ref{ccr algorithm}, with the only difference being the update steps in line 11 and 20, where the
    TD error must be calculated and sent to the optimizer to perform backpropagation. Additionally, before calculating the TD error, the experiences containing the CCR must be
    stored in the experience replay memory. A similar methodology is used when applying CCR to deep recurrent Q-learning, and will only affect the experiences stored within the
    short term memory.

\section{A quintessential example}\label{setting_1}
    We test independent Q-learning and its CCR variant within a simplified scenario with a small state space to show the fundamental limitation of RL methods, even for the tabular
    case. We will then show how our variation improves upon them. We will discuss the experimental design and briefly highlight the main results. 

    The environment is a cooperative two-player card game. Each player receives 3 cards with rank 1, 2, or 3, but cannot see their own cards, they can only see the cards of the
    other player. In the center is a target card visible to both players, and the aim of the game is to play a card matching the target card. On their turn a player can, either
    play a card from their hand or hint a card in the other player's hand. The game ends as soon as a card is played, and if the card matched the target, the play was successful
    and the players receive a score of 1. Otherwise, the players loose with a score of 0. The solution to this problem is trivial and only requires two turns. The first player must
    hint to the other player which card matches the target, and on the next turn the player must play that card. For a discussion on the hyperparameters used in each method see
    Table~\ref{table of hypers} in Appendix A.

\begin{figure}[h]
        \centerline{\includegraphics[width=21pc]{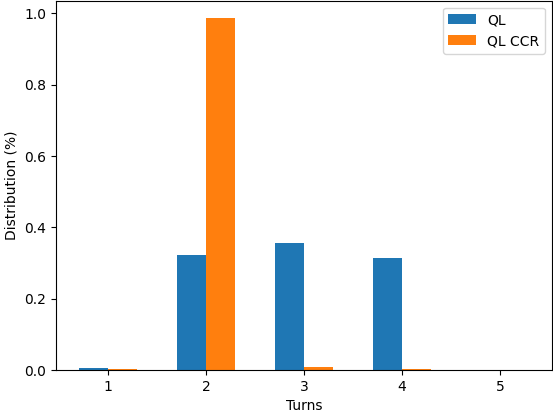}} \caption{Distribution of the number of steps taken by independent Q-learning (QL) and independent
        Q-learning with CCR applied (QL CCR) agents over the course of 1 000 evaluation episodes, after training for 100 000 episodes.}\label{lr_curves_ql_ccr}
    \end{figure}

    Fig.~\ref{lr_curves_ql} shows the learning curves for each method over the course of 25 000 episodes. It is clear that both methods are capable of solving the problem and
    achieve a perfect score of 1, but independent Q-learning requires a considerable number of training steps to do so. Therefore, it is clear that independent Q-learning agents
    struggles to learn in this simplified problem. We can further justify this statement by plotting the number of steps before the game ended in perfect score, shown in
    Fig.~\ref{lr_curves_ql_ccr}. From these results, we can see that, although the independent Q-learning agents are able to solve the problem, they cannot learn the optimal
    policy. In contrast, the CCR agents display significantly faster training capabilities and converges at the perfect score with the optimal policy. 

    \begin{figure}[h]
        \centerline{\includegraphics[width=21pc]{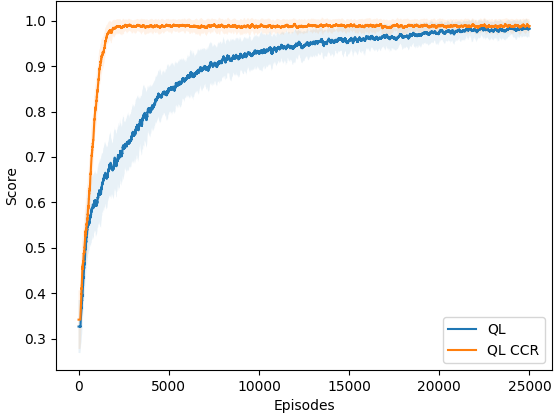}} \caption{Learning curves for independent Q-learning (QL) and independent Q-learning with CCR
        applied (QL CCR) with a 100-episodes moving average and standard deviation, averaged over 50 runs per method.}\label{lr_curves_ql}
    \end{figure}

\section{Performance Evaluation using Colourless Hanabi}\label{setting_2} 
    To test and evaluate the performance of CCRs when applied to deep RL, we use a simplified version of the cooperative card game Hanabi, called colourless Hanabi. Bard \textit{et
    al.}~\cite{hanabi_ai} propose Hanabi as a new frontier for the testing and development of cooperative RL algorithms. It offers an intricate challenge testing various concepts
    crucial for successful cooperation. 
    
    Using colourless Hanabi we will evaluate the performance of deep Q-learning, n-step deep Q-learning, CCR applied to deep Q-learning, deep recurrent Q-learning and CCR applied
    to deep recurrent Q-learning. All these methods' hyperparameters are optimized to ensure the comparison of each method's best agents. We will first introduce colourless Hanabi
    and discuss the experimental setup for each method. This is followed by the results for each method's performance and an evaluation of the learned policies. 

\subsection{Colourless Hanabi}
    Colourless Hanabi is a cooperative two-player card game. The aim of the game is to consecutively stack cards in numerical order from one to five. The deck comprises of 20 cards
    with a distribution of six 1s, four 2s, four 3s, four 4s, and two 5s. At the start of the game, each player receives five cards randomly from the deck. Similar to the tabular
    problem, a player cannot see their own cards and can only see those of the other player. 
    
\begin{figure}[h]
        \centerline{\includegraphics[width=21pc]{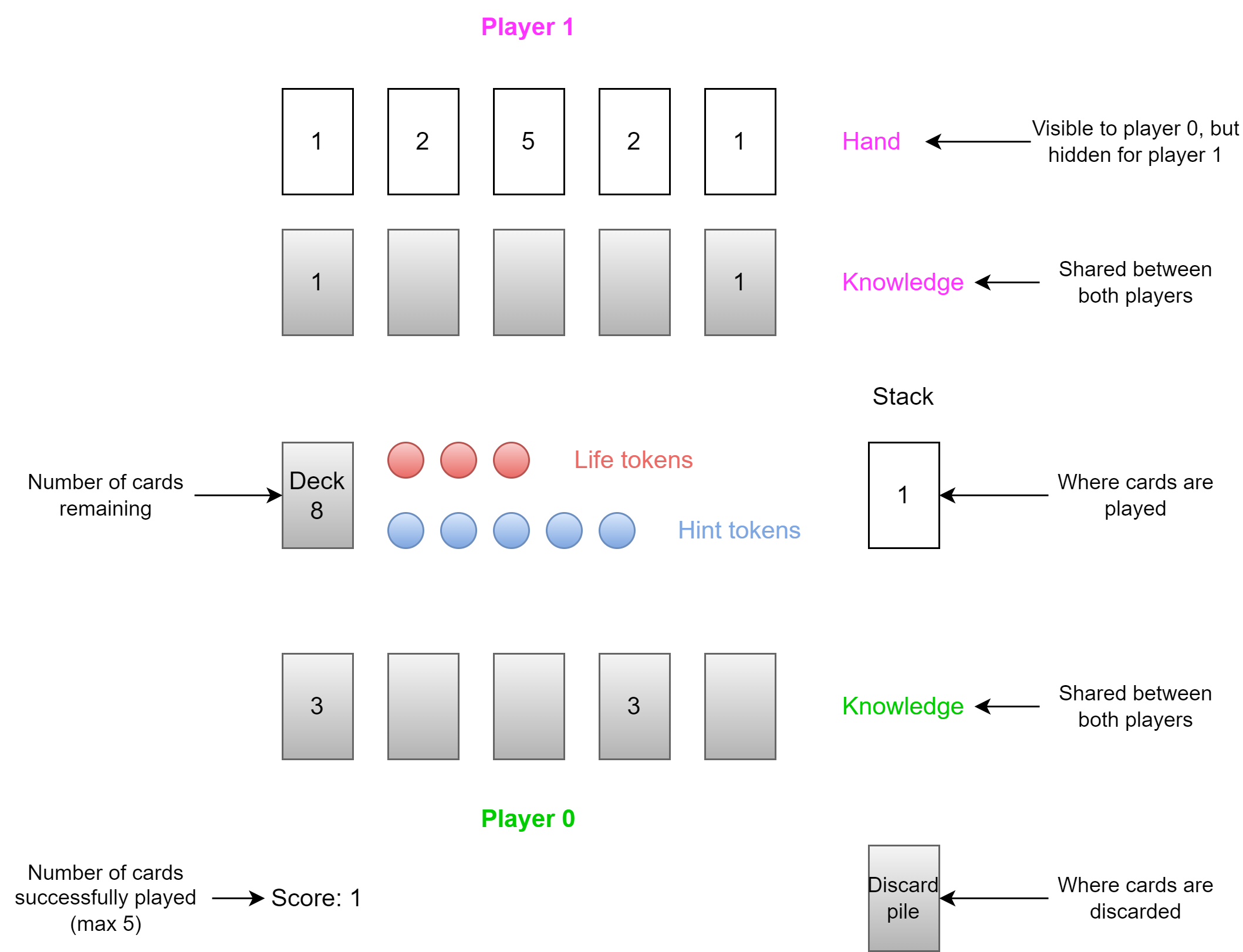}} \caption{An example game of colourless Hanabi as seen from the perspective of player 0. Player 1
        has already hinted two 3s in player 0's hand, similarly player 0 has hinted two 1s in player 1's hand. The deck has 8 cards remaining, and the players have all three life
        tokens as well as six hint tokens remaining. One card has already been placed on the stack and the players, therefore, have a score of 1/5. Player 0 can now take the hint,
        play or discard action.}\label{colourless_han}
    \end{figure}

    The players take consecutive turns, and on each turn a player can either \emph{play}, \emph{discard}, or \emph{hint}. Hinting involves revealing all the cards in the other
    player's hand matching a certain rank. In the centre is a stack where the players must play their cards, and a successful play entails playing a card that follows the current
    card on top of the stack (starting at 0). If the card played does not follow the current card on the stack, the play was unsuccessful (called a \emph{misplay}) and the
    players lose one of their shared life tokens. An example game is shown in Fig.~\ref{colourless_han}. 

    At the start of the game the players have a shared total of three life tokens and eight hint tokens. If all the hint tokens are depleted, a player cannot take the hint action
    and must either play or discard a card from their hand. The players are awarded a shared score depending on the number of cards successfully placed on the stack. If the players
    manage to build the stack to the maximum of 5, the game ends and the players receive a perfect score of 5/5. Alternatively, if the players lose all their life tokens, or if the
    deck is depleted, the game ends. A discard action, or a misplay, will result in the card being removed from the game. When a player choses the discard or play action, they draw
    a new card from the deck to replenish the missing card (which is also hidden from current the player). This problem requires a delicate action sequence to solve. Even if the
    first action is optimal, it amounts to nothing without the appropriate response. A player requires knowledge that only the other player can give in order to win.  

\subsection{Experimental Setup}
\begin{figure*}[!t]
        \centerline{\includegraphics[width=42pc]{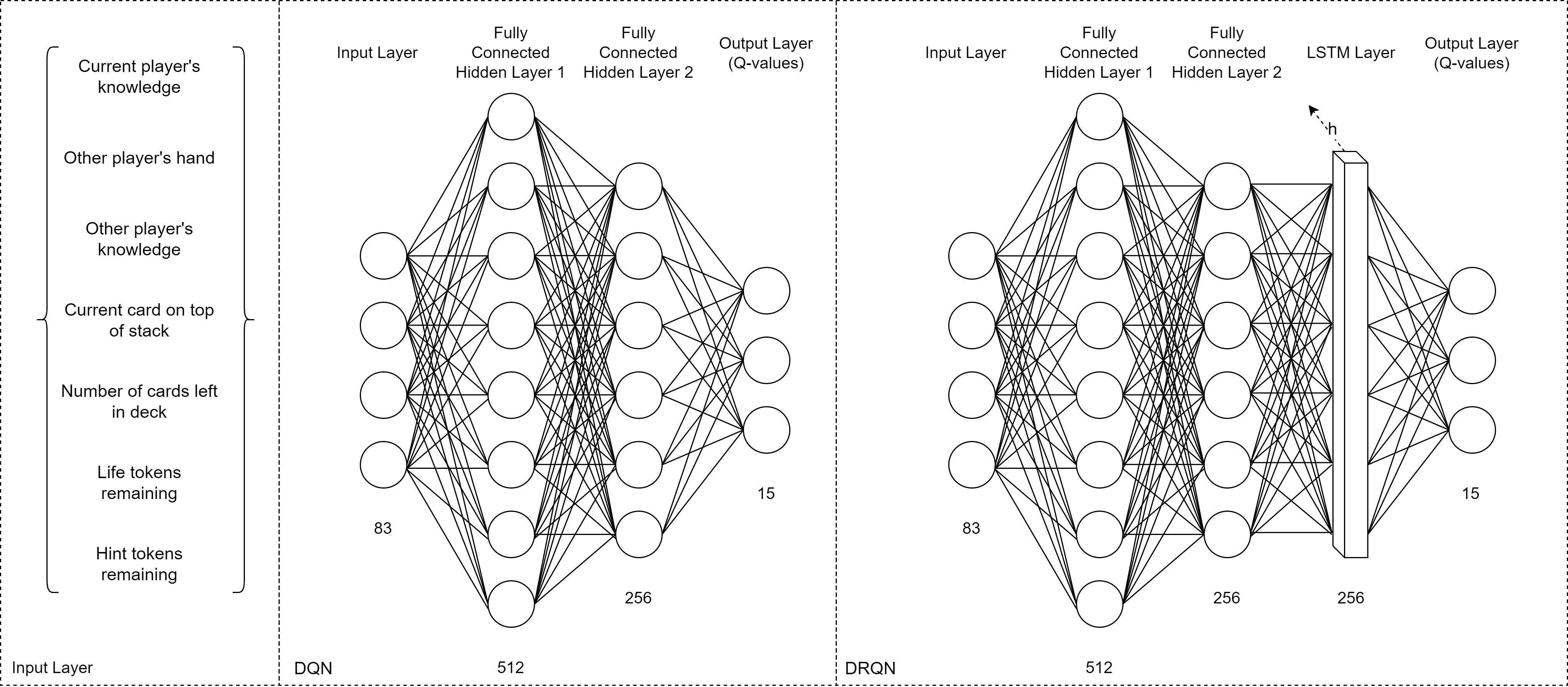}} \caption{Architecture design for deep Q-learning (DQN) and deep recurrent Q-learning (DRQN), along
        with the structure of the input layer. The number of \emph{neurons} (or units) are shown below each layer. In the case of DRQN, the weights $h$ of the recurrent layer are
        passed between episodes to form the histories.}\label{arch}
    \end{figure*}

    Before discussing the results, we will briefly highlight the architecture design for each deep RL technique. For a full list of the hyperparameters used in each method see
    Table~\ref{table of hypers} in Appendix A. All the methods received a one-hot encoded tuple shown in Fig.~\ref{arch} as input to their neural networks.

    The deep Q-learning and n-step deep Q-learning solutions use a feed forward neural network as shown in Fig.~\ref{arch}, utilising the \textit{ReLU} activation
    functions~\cite{relu}. Both methods implement the Adam optimizer~\cite{adam_opt} to calculate the TD error and perform backpropagations to update the network weights. Deep
    recurrent Q-learning uses a similar architecture as deep Q-learning with two hidden feed forward neural networks, but with the addition of a LSTM recurrent layer before the
    outputs. It also uses the Adam optimizer to calculate the TD error and implements an unrolled length of 2.

    When applying CCR to deep Q-learning and deep recurrent Q-learning, we only change the transition tuples stored within each method's memory. Although this does not influence
    the network architecture, the optimal hyperparameters are different, as shown in Table~\ref{table of hypers} in Appendix A. During evaluation we compare each method's 100
    episode moving averages and standard deviations, averaged over 20 different runs. 

    Due to the symmetric nature of the environment, we apply a shared policy strategy for independent deep Q-learning, deep recurrent Q-learning and their CCR variants. As a
    result, the agents act as copies of each other, similar to a self-play scenario~\cite{self_play}. However, this still restricts learning by only using individual experiences,
    i.e., there is no additional sharing of state information between each agent, the agents merely share an action-value function. During evaluation the agents are completely
    separate, and do not share any additional information. 

\subsection{Performance of CCRs}\label{results}
\begin{figure}[h]
        \centerline{\includegraphics[width=21pc]{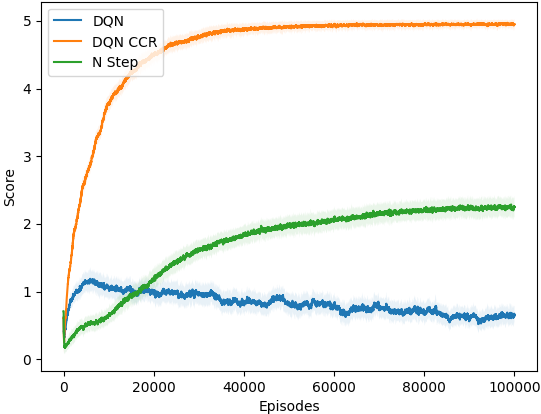}} \caption{Learning curves for deep Q-learning (DQN), n-step deep Q-learning (N Step), and CCR
        applied to deep Q-learning (DQN CCR) with a 100-episode moving average and standard deviation, averaged over 20 runs per method.}\label{lr_curves_dqn}
    \end{figure}

    The learning curves for DQN, n-step DQN, and DQN-CCR over the course of 100 000 episodes are shown in Fig.~\ref{lr_curves_dqn}. This shows that the DQN agents display an
    inability to learn within the example setting. The average score achieved never exceeds 1/5 and starts to deteriorate as training continues, with the agents devolving to
    semi-random behaviour, further demonstrating the agents' inability to learn delicate action sequences. If instead we use n-step bootstrapping, the performance increases
    slightly, but the agents still perform inadequately. 

    DQN-CCR significantly improves the performance of the DQN agents. The agents are able to learn and achieve a near perfect score of 5/5, indicating that they are able to solve
    the problem. The CCR variant displays a significantly faster steady-state response and improves on the performance of n-step bootstrapping. This illustrates that by
    incorporating the reward of the other agents into an agent's return, we are able to encourage the development of delicate action sequences, crucial for solving cooperative tasks. 

\begin{figure}[h]
        \centerline{\includegraphics[width=21pc]{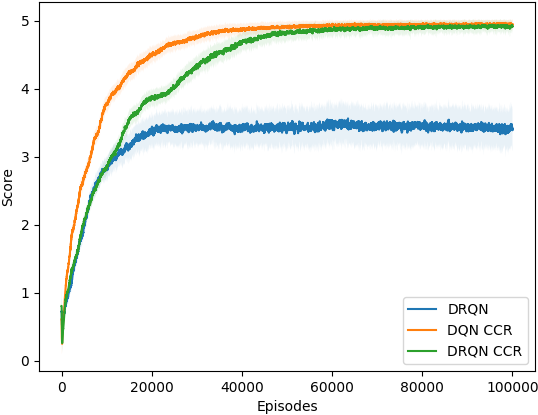}} \caption{Learning curves for deep recurrent Q-learning (DRQN), CCR applied to deep
        Q-learning (DQN CCR), and CCR applied to deep recurrent Q-learning (DRQN CCR) with a 100-episode moving average and standard deviation, averaged over 20 runs per
        method.}\label{lr_curves_drqn}
    \end{figure}

    We continued our investigation and applied recurrent networks to see if it offers any performance gain. Fig.~\ref{lr_curves_drqn} shows the learning curves for DRQN and
    DRQN-CCR, and compare them to DQN-CCR (from Fig.~\ref{lr_curves_dqn}) as a baseline. It is evident that the DRQN agents are able to improve on the performance of the DQN
    agents, as well as n-step bootstrapping, however, it is still not optimal. DRQN displays a fast steady-state response when compared to n-step bootstrapping, but with a
    significant amount of standard deviation. This indicates instability within the learnt policies. By applying CCR to DRQN, we are able to further improve on its performance and
    mitigate its shortcomings. The DRQN-CCR agents are able to achieve a perfect score and display less deviation within the convergent policy. The DQN-CCR agents still displays
    the best performance, but after enough training episodes, the DRQN-CCR agents achieve a similar steady-state response. 

    These results show that the DQN agents are not able to learn within this environment and by using recurrent networks, we are able to improve on the performance. However, the agents
    are not able to achieve a perfect score and therefore do not solve the problem. The results show that by applying CCRs to both cases, we are able to significantly improve on
    these shortcomings and the agents are able to solve the problem. From the learning curves in Fig.~\ref{lr_curves_drqn}, the DQN-CCR agents perform the best of all the methods.
    It converges (a) the fastest, and (b) to the perfect score. It also displays less standard deviation, indicating a more stable policy. 

\subsection{Learned Policies and Conventions}
    We next evaluate the learned policies of each method after training for 100 000 episodes. The results are shown in Table~\ref{table for val comps}. The oracle is a rule-based
    agent that is handcrafted to be able to solve colourless Hanabi for the best case scenario, where the deck is perfectly distributed and the need for discarding thereby omitted.
    This will allow the game to be ended within 10 steps, since a player hints a card and on the next turn that card is played.

    \begin{table*} [!t]
        \caption{Evaluation of each method's policy over the course of 1 000 episodes. The best result for each category compared to the oracle are shown in bold.}\label{table for val comps}
        \setlength{\tabcolsep}{3pt}
        \begin{tabular*}{42pc}{@{}|p{50pt}|p{25pt}<{\raggedright\hangindent6pt}|p{50pt}<{\raggedright\hangindent6pt}|p{25pt}<{\raggedright\hangindent6pt}|p{25pt}<{\raggedright\hangindent6pt}|p{35pt}<{\raggedright\hangindent6pt}|p{35pt}<{\raggedright\hangindent6pt}|p{95pt}<{\raggedright\hangindent6pt}|p{100pt}<{\raggedright\hangindent6pt}|@{}}
            \hline
            \textbf{Method}    &   \textbf{Score}  &    \textbf{Total Actions}  & \textbf{Hints}    &   \textbf{Plays}   &   \textbf{Misplays (\%)}    &   \textbf{Discards (\%)}
            & \textbf{Average steps to perfect score}  &   \textbf{Games ending in perfect score (\%)}    \\
            \hline
            Oracle          &   5               &   10 000          &     5 000         &   5 000           &   0.00~\%             &    0.00~\%            &   10.00           &   100~\%              \\
            \hline
            DQN             &   0.882           &   12 657          &     1 758         &   1 305           &   3.34~\%             &   75.80~\%            &   None            &   0.00~\%             \\
            \hline
            N Step DQN      &   2.325           &   14 551          &     3 623         &   2 508           &   1.26~\%             &   57.87~\%            &   None            &   0.00~\%             \\
            \hline
            DQN CCR         &   \textbf{4.975}  &  11 750           &   \textbf{5 451}  &   \textbf{5 008}  &   \textbf{0.28~\%}    &   \textbf{10.99~\%}   &   11.58           &   \textbf{98.10~\%}   \\
            \hline
            DRQN            &   3.556           &   9 898           &     1 219         &   5 926           &   25.18~\%            &   27.81~\%            &   11.82           &   19.70~\%            \\
            \hline
            DRQN CCR        &   4.938           &   \textbf{9 946}  &     3 524         &   5 222           &   2.86~\%             &   12.07~\             &   \textbf{9.78}   &   95.70~\%            \\
            \hline
            \multicolumn{9}{@{}p{42pc}@{}}{\hspace*{9pt}The number of misplays and discards are taken as percentages over the total number of actions.}\\
        \end{tabular*}\label{tab2}
    \end{table*}

    From these results we can see that the DQN and n-step DQN agents show the worst performance, with the lowest scores, least number of plays, most discards and the inability to
    achieve a perfect score. The DRQN agents manage to achieve a higher average score compared to the DQN and n-step DQN agents, but with a high percentage of misplays and the
    lowest number of hints. When monitoring the DRQN agents' behaviour we observe a high risk policy, where the agents often take the play action at random and only start to act
    conservatively when the number of life tokens are low. This, however, is not an ideal policy since it can result in important cards being discarded from the game, and is
    supported by the agents only managing a success rate of 19.7\%. The DQN-CCR and DRQN-CCR agents respectively manage to achieve a perfect score 98.1\% and 95.7\% of the time.
    Both methods achieve similar scores, number of plays and percentage of discards indicating that they are able to solve the problem.

    We argue that the presented results show that current RL formulations struggle to develop delicate action sequences when they are unable to observe the effect their actions had
    on the environment, and on other agents. Our results show that by using our variation on the transition tuple for turn-based MARL problems, we are able to significantly improve
    on the performance of independent Q-learning, independent deep Q-learning and deep recurrent Q-learning. We also illustrate the difference between our novel CCR variant and
    n-step bootstrapping, which shares similarities, but implements a different concept and produces different results. 

\section{Conclusion}\label{concl}
    We proved that delicate action sequences pose a challenge to existing MARL formulations. Without observing the effect an agent's action had on the environment as well as its
    co-agents, can lead to inadequate policy development and inability to cooperate. We proposed a new variant, called credit-cognisant rewards, which addresses these challenges and
    encourages the development of delicate action sequences by incorporating all the rewards within an action sequence into the immediate reward of an individual agent's action. 

    We demonstrated how our CCR variant improved on the performance of independent Q-learning, by producing a faster training time and superior convergent policy. We
    extended this to colourless Hanabi, which is a simplified version of the cooperative card game Hanabi, and explored the capabilities of independent deep Q-learning and deep
    recurrent Q-learning. This further justified the shortcomings of existing MARL techniques, and we showed how our CCR variant continues to improve upon them. CCR applied to deep
    Q-learning produced the best overall performance, with the highest average score and success rate. It also displayed the lowest percentage of misplays and the agents achieved
    the closest results to the oracle. 
    
    Future work can further explore the capabilities of CCRs, and apply it to other value-based and or policy gradient methods. This could offer additional performance gain and
    allow for added capabilities of MARL agents, with the end goal of solving more complex problems requiring agent cooperation. 

\section*{Appendix A: Hyperparameters}\label{hyperparams}
    Herein follows a list of the hyperparameters used for independent Q-learning, deep Q-learning, n-step deep Q-learning, deep recurrent Q-learning, and their CCR variants shown
    in Table~\ref{table of hypers}. Each method's hyperparameters were optimized using parameter sweeps, i.e., training each method with various combinations of hyperparameters
    over the course of multiple runs and selecting the hyperparameters with the best results.
    
    \begin{table*} [!t]
        \caption{Table of hyperparameter values for each method used in our evaluations. Each hyperparameter sweep were conducted over 10 runs consisting of 50 000
        episodes.}\label{table of hypers}
        \setlength{\tabcolsep}{3pt}
        \centering
        \begin{tabular*}{42pc}{@{}|p{115pt}|p{23pt}<{\raggedright\hangindent6pt}|p{345pt}<{\raggedright\hangindent6pt}|@{}}
            \hline
            \multicolumn{3}{|c|}{\textbf{Independent Q-learning}} \\
            \hline
            Hyperparameter                  &   Value   &   Description \\
            \hline
            Learning rate $\alpha$          &   0.1     &   Learning rate used in Q-learning update step \\
            Discount factor $\gamma$        &   0.9     &   Discount factor used in Q-learning update step \\
            Exploration $\epsilon$          &   0.01    &   Value of $\epsilon$ in $\epsilon$-greedy strategy \\
            \hline
            \multicolumn{3}{|c|}{\textbf{Independent Q-learning with CCR}} \\
            \hline
            Hyperparameter                  &   Value   &   Description \\
            \hline
            Learning rate $\alpha$          &   0.01    &   Learning rate used in Q-learning update step \\
            Discount factor $\gamma$        &   0.5     &   Discount factor used in Q-learning update step \\
            Exploration $\epsilon$          &   0.01    &   Value of $\epsilon$ in $\epsilon$-greedy strategy \\
            \hline
            \multicolumn{3}{|c|}{\textbf{Deep Q-learning}} \\
            \hline
            Hyperparameter                  &   Value   &   Description \\
            \hline
            Learning rate $\alpha$          &   0.0001  &   Learning rate used by Adam optimizer \\
            Discount factor $\gamma$        &   0.7     &   Discount factor used in Q-learning update step \\
            Exploration $\epsilon$          &   0.01    &   Value of $\epsilon$ in $\epsilon$-greedy strategy \\
            Replay memory size              &   10 000  &   Where experience are stored to use in the update step \\
            Sampling batch size             &   64      &   Number of experiences randomly sampled from the experience replay memory used in the update step \\
            Target network update frequency &   100     &   Number of steps before the target network is updated with the policy network \\
            \hline
            \multicolumn{3}{|c|}{\textbf{N-Step Deep Q-learning}} \\
            \hline
            Hyperparameter                  &   Value   &   Description \\
            \hline
            Learning rate $\alpha$          &   0.0001  &   Learning rate used by Adam optimizer \\
            Discount factor $\gamma$        &   0.3     &   Discount factor used in Q-learning update step \\
            Exploration $\epsilon$          &   0.01    &   Value of $\epsilon$ in $\epsilon$-greedy strategy \\
            Replay memory size              &   10 000  &   Where experience are stored to use in the update step \\
            Sampling batch size             &   64      &   Number of experiences randomly sampled from the experience replay memory used in the update step \\
            Target network update frequency &   100     &   Number of steps before the target network is updated with the policy network \\
            Step count $n$                  &   2       &   Number of steps over which the return is constructed \\
            \hline
            \multicolumn{3}{|c|}{\textbf{Deep Q-learning with CCR}} \\
            \hline
            Hyperparameter                  &   Value   &   Description \\
            \hline
            Learning rate $\alpha$          &   0.0001  &   Learning rate used by Adam optimizer \\
            Discount factor $\gamma$        &   0.5     &   Discount factor used in Q-learning update step \\
            Exploration $\epsilon$          &   0.01    &   Value of $\epsilon$ in $\epsilon$-greedy strategy \\
            Replay memory size              &   10 000  &   Where experience are stored to use in the update step \\
            Sampling batch size             &   64      &   Number of experiences randomly sampled from the experience replay memory used in the update step \\
            Target network update frequency &   100     &   Number of steps before the target network is updated with the policy network \\
            \hline
            \multicolumn{3}{|c|}{\textbf{Deep Recurrent Q-learning}} \\
            \hline
            Hyperparameter                  &   Value   &   Description \\
            \hline
            Learning rate $\alpha$          &   0.0001  &   Learning rate used by Adam optimizer \\
            Discount factor $\gamma$        &   0.5     &   Discount factor used in Q-learning update step \\
            Exploration $\epsilon$          &   0.01    &   Value of $\epsilon$ in $\epsilon$-greedy strategy \\
            Short term memory size          &   5 000   &   Where experience are stored in the hidden layer to use in the update step \\
            Sampling batch size             &   32      &   Number of experiences randomly sampled from the experience replay memory used in the update step \\
            Target network update frequency &   100     &   Number of steps before the target network is updated with the policy network \\
            Recurrent layer types           &   LSTM    &   Type of recurrent layer used in the network architecture \\
            Recurrent layer count           &   1       &   Number of recurrent layers before the output \\
            Unrolled length $K$             &   2       &   Length of the unrolled experiences to use in the update step  \\
            Maximum episode length          &   50      &   Maximum number of steps for a episode to be stored in memory \\
            \hline
            \multicolumn{3}{|c|}{\textbf{Deep Recurrent Q-learning with CCR}} \\
            \hline
            Hyperparameter                  &   Value   &   Description \\
            \hline
            Learning rate $\alpha$          &   0.0001  &   Learning rate used by Adam optimizer \\
            Discount factor $\gamma$        &   0.1     &   Discount factor used in Q-learning update step \\
            Exploration $\epsilon$          &   0.01    &   Value of $\epsilon$ in $\epsilon$-greedy strategy \\
            Short term memory size          &   5 000   &   Where experience are stored in the hidden layer to use in the update step \\
            Sampling batch size             &   32      &   Number of experiences randomly sampled from the experience replay memory used in the update step \\
            Target network update frequency &   100     &   Number of steps before the target network is updated with the policy network \\
            Recurrent layer types           &   LSTM    &   Type of recurrent layer used in the network architecture \\
            Recurrent layer count           &   1       &   Number of recurrent layers before the output \\
            Unrolled length $K$             &   2       &   Length of the unrolled experiences to use in the update step  \\
            Maximum episode length          &   50      &   Maximum number of steps for a episode to be stored in memory \\
            \hline
            \multicolumn{3}{@{}p{42pc}@{}}{}\\
        \end{tabular*}\label{tab5}
    \end{table*} 

\section*{Acknowledgment}
    Computations were performed using Stellenbosch University's HPC1 (\href{http://www.sun.ac.za/hpc}{Rhasatsha}).

\bibliography{brede_refs.bib}

\end{document}